\documentclass[letterpaper, 10 pt, conference]{ieeeconf}  

\usepackage{amsmath}
\usepackage{verbatim}
\usepackage{graphicx}
\usepackage{gensymb}
\usepackage{color,soul}
\usepackage[font={small,it}]{caption}
\usepackage{subcaption}
\usepackage{rotating}
\usepackage{multirow}
\usepackage{tikz}
\usetikzlibrary{arrows,shapes}

\renewcommand{\baselinestretch}{0.95}

\title{Visual Closed-Loop Control for Pouring Liquids}

\author{Connor Schenck, Dieter Fox\\University of Washington\thanks{{\bf Acknowledgement} This work was funded in part by the National Science Foundation under contract numbers NSF-NRI-1525251 and NSF-NRI-1637479.}}

\IEEEoverridecommandlockouts
\begin{document}

\maketitle
\thispagestyle{empty}
\pagestyle{empty}

\begin{abstract}

Pouring a specific amount of liquid is a challenging task.
In this paper we develop methods for robots to use visual feedback to perform closed-loop control for pouring liquids.
We propose both a model-based and a model-free method utilizing deep learning for estimating the volume of liquid in a container.
Our results show that the model-free method is better able to estimate the volume.
We combine this with a simple PID controller to pour specific amounts of liquid, and show that the robot is able to achieve an average 38ml deviation from the target amount.
To our knowledge, this is the first use of raw visual feedback to pour liquids in robotics.

\end{abstract}

\section{Introduction}

The last years have seen dramatic improvements in robotic capabilities relevant
to household tasks such as putting items into a
dishwasher~\cite{jiang2012}, folding and ironing
clothing~\cite{miller2012,li2016multi}, and cleaning
surfaces~\cite{xu2014}.  So far, however, robots have not been able to
robustly perform household tasks involving liquids, such as pouring a glass of
water.  Solving such tasks requires both robust control and detection of liquid
during the pouring operation. 
Humans often are not very accurate at this, requiring specialized containers to measure a specific amount of liquid.
Instead, people often use vague, relative terms such as ``Pour me a half cup of coffee'' or ``Just a little, please.''
While there has been recent success in robotics
on controlling a manipulator to pour liquids simulated by small
balls~\cite{yamaguchi2015} and on detecting liquids using optical flow or deep
learning~\cite{yamaguchi2016,schenckc2016b}, the task of pouring certain amounts of actual
liquids has not been addressed.

In this paper, we introduce a framework that enables robots to robustly pour
specific amounts of a liquid into containers typically found in a home
environment, such as coffee mugs, cups, glasses, or bowls.  We achieve
this in the most general setting, without requiring specialized hardware, such
as highly accurate force sensors for measuring the amount of liquid held by a
robot manipulator, scales placed under the target container, or sensors designed for
detecting liquids.
However, while we avoid requiring specialized environmental augmentation, our investigation is on how accurate a robot could pour under relatively controlled conditions, such as having been able to train on the target containers.

The intuition behind our approach is based on the insight that people strongly
rely on visual cues when pouring liquids.  For example, a health study revealed
that the amount of wine people pour into a glass is strongly biased by visual
factors such as the shape of the glass or the color of the wine~\cite{Wal14Hal}.
We thus propose a framework that uses visual feedback in a closed-loop pouring
controller.  Specifically, we train a deep neural network structure to estimate
the amount of liquid in a cup from raw visual data.  Our network structure has
two stages. In the first stage, a network detects which pixels in a camera image
contain water.  The output of the detection network is fed into another network
that estimates the amount of liquid already in the container. This amount is used as
real-time feedback in a PID controller that is tasked to pour a desired amount
of water into a cup.

To generate labeled data needed for the neural networks, we developed an
experimental setup that uses a thermal camera calibrated with an RGBD camera to
automatically label which pixels in the color frames contain (heated) water.
Experiments with a Baxter robot pouring water into three different containers
(two mugs and one bowl) indicate that this approach allows us to train deep
networks that provide sufficiently accurate volume estimates for the pouring
task.

\begin{figure}[t]
    \centering
    \setlength{\fboxsep}{0pt}
    \setlength{\fboxrule}{1pt}
    \setlength{\unitlength}{1.0cm}
    \fbox{\includegraphics[width=5.0cm]{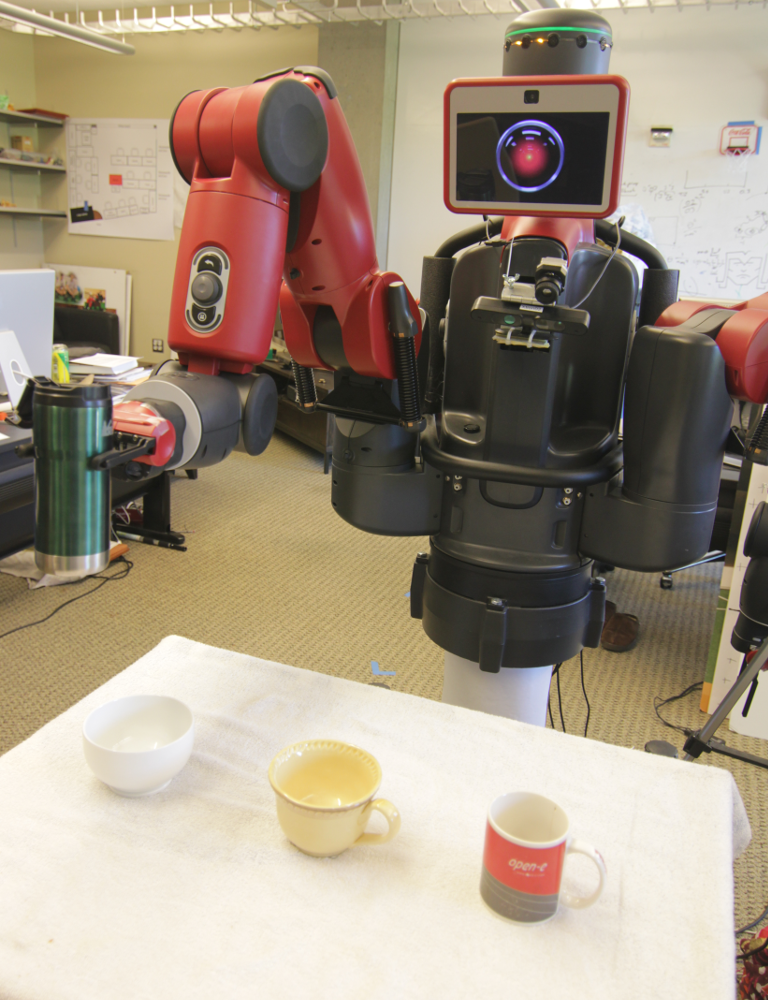}}
    \caption{The Baxter robot used in our experiments. In its right gripper it holds the cup used as the source container. On the table in front of the robot are the three target containers (from left to right): the small bowl, tan mug, and redgray mug.}
    \label{fig:robot_setup}
    \vspace{-0.75cm}
\end{figure}

Our main contributions in this paper are (1) an overall framework for
determining the amount of liquid in a container for real-time control during a
pouring action; (2) the use of thermal imagery to generate ground truth data for
pixel level labeling of (heated) water; (3) a deep neural network that uses such
labels to detect liquid pixels in raw color images; (4) a model-based method to
determine the volume of liquid in a target container given pixel-wise liquid
detection; (5) a neural network to regress to the volume of liquid given
pixel-wise liquid detections as input; and (6) an extensive evaluation that
shows that our methodology is suitable for control by deploying it on a robot
for use in a pouring task.

\section{Related Work} 

\begin{figure*}[t]
    \centering
    \setlength{\fboxsep}{0pt}
    \setlength{\fboxrule}{1pt}
    \setlength{\unitlength}{1.0cm}
    \vspace{0.2cm}
    \begin{subfigure}{3.5cm}
        \fbox{\includegraphics[width=3.5cm]{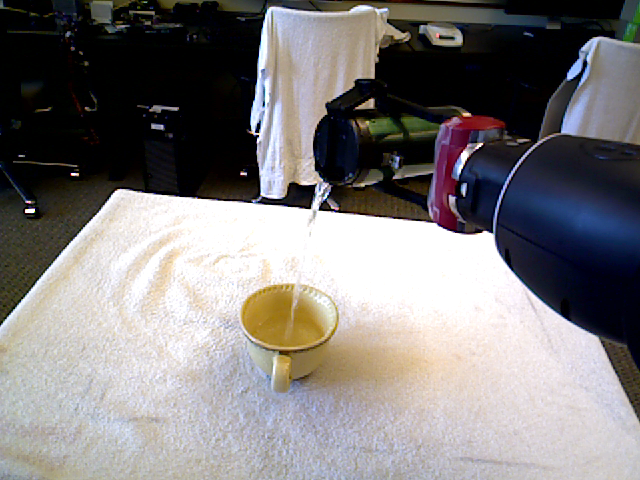}}
        \caption{RGB}
        \label{fig:rgb_example}
    \end{subfigure}\hspace{0.2cm}%
    \begin{subfigure}{3.5cm}
        \fbox{\includegraphics[width=3.5cm]{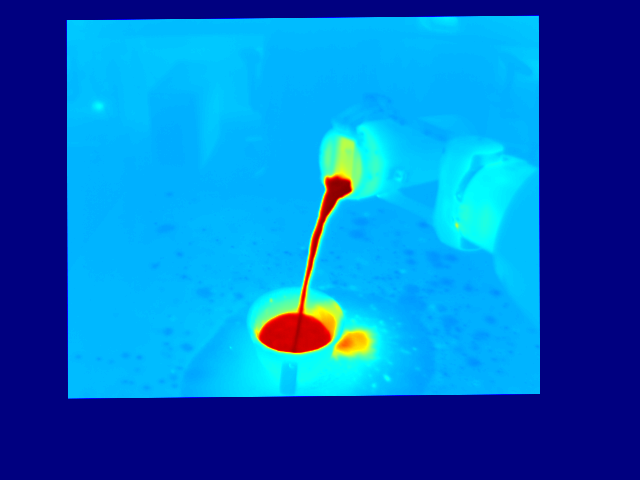}}
        \caption{Thermal}
        \label{fig:thermal_example}
    \end{subfigure}\hspace{0.2cm}%
    \begin{subfigure}{3.5cm}
        \fbox{\includegraphics[width=3.5cm]{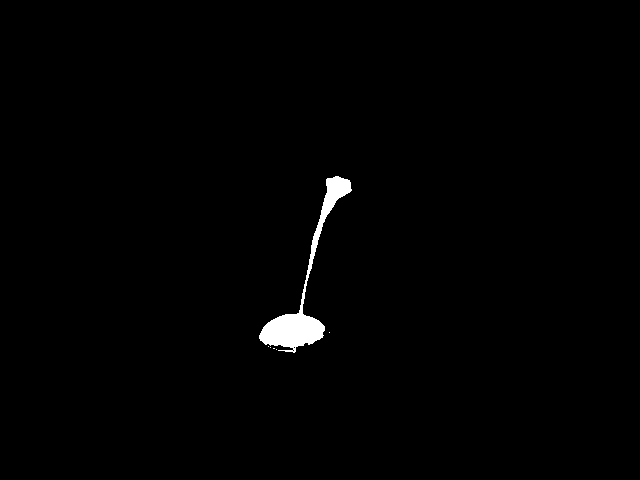}}
        \caption{Thresholded thermal}
        \label{fig:threshold_example}
    \end{subfigure}\hspace{0.2cm}%
    \begin{subfigure}{3.5cm}
        \fbox{\includegraphics[width=3.5cm]{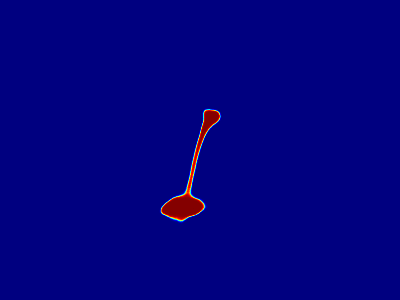}}
        \caption{Network detections}
        \label{fig:detection_example}
    \end{subfigure}
    \vspace{-0.15cm}
    \caption{An example frame from a pouring sequence. The left image shows the
      color camera from the robot's perspective, the left-middle image shows a
      heatmap of the thermal camera after it has been registered to the color
      camera, the right-middle image shows the water labeled by thresholding the
      thermal image, and the right image shows the output of the detection
      network based on the RGB input.}
    \label{fig:therm_examples}
    \vspace{-0.75cm}
\end{figure*}

There is prior work related to robotic pouring, however, most of it either uses coarse simulations disconnected from real liquid perception and dynamics \cite{kunze2015,yamaguchi2015} or constrained task spaces that bypass the need to perceive and reason directly about liquids \cite{langsfeld2014,okada2006,tamosiunaite2011,cakmak2012,rozo2013}.
Additionally, all of these works with the exception of \cite{rozo2013} pour the entire contents of the source container into the target container, with the focus on other factors such as spillage or the overall motion trajectory.
In contrast, in this work we focus primarily on pouring a specific amount of liquid from the source into the target rather than simply emptying the source container into the target.
To do this, the robot requires some method for estimating the volume of liquid in the target.
Rozo {\it et al.}\cite{rozo2013} utilized force sensors in the robot's arm to measure how much had been poured out, however this requires a robot with very precise torque sensors, which are not available on our Baxter robot.
In our own prior work \cite{schenckc2016a} we placed a digital scale under the target container.
But this method presents many of its own challenges, such as delay in the scale measurement (often 1-2 seconds) and no information about where the liquid is or how it is moving.
Humans, on the other hand, are able to accomplish this task purely from visual feedback, which strongly suggests that robots should be able to as well.

There is some prior work related to directly perceiving liquids from sensory feedback \cite{griffith2012,rankin2010,rankin2011}.
Work by Yamaguchi and Atkeson \cite{yamaguchi2016} utilizes optical flow to detect liquids as they flow from a source into a target container.
However, for the tasks in this paper, the robot must also be able to detect standing water with no motion, for which optical flow is poorly suited.
Instead, we build on our own prior work relating to liquid detection in simulation \cite{schenckc2016b}.
We developed a method utilizing fully-convolutional neural networks \cite{long2015} to label pixels in an image as either {\it liquid} or {\it not-liquid}.
Here we utilize the recurrent network with long short-term memory (LSTM) layers \cite{hochreiter1997} that we used in that work to detect and label liquid in an image.



\section{Technical Approach}

\subsection{Task Overview}
In this paper, the robot is tasked with pouring a specific amount of liquid from a source container into a target container.
This task is more difficult than prior work on robotic pouring which primarily focuses on pouring all the contents of the source container into the target container, whereas we focus on pouring only a limited amount from the source containing an unkown initial amount of liquid.
To accomplish this, the robot must use visual feedback to continuously estimate the current volume of liquid in the target container.
Our approach has 3 main components:
First the robot detects which pixels in its visual field are liquid and which are not.
Next the robot uses these detections to estimate the volume of liquid in the target container.
Finally, the robot feeds these volume estimates into a controller to pour the liquid into the target.
Figure \ref{fig:model} shows a diagram of this process.
We structure the problem in this manner as opposed to simply training one end-to-end network as it allows us to train and evaluate each of the individual components of the system, which can give us better insight into its operation.

\subsection{Pixel-Wise Liquid Detection}
\label{sec:classification}

In order for both the model-based and model-free volume estimation methods to work, the robot must classify each pixel in the image as {\it liquid} or {\it not-liquid}. 
We developed two methods for acquiring these pixel labels: a thermographic camera in conjunction with heated water, and a fully-convolutional neural network\cite{long2015} with color images.
While the thermal camera works well for generating pixel labels, it is also rather expensive and must be registered to an RGBD sensor. 
In our prior work \cite{schenckc2016b}, we developed a method for generating pixel labels on simulated data for liquid from color images only, which we briefly describe here.

\begin{figure*}
\begin{center}
\vspace{0.5cm}
\begin{tikzpicture}[->,>=stealth,auto,node distance=1.6cm,thick,
      input node/.style={rectangle,draw,anchor=west,align=center,inner sep=0,outer sep=0},
      output node/.style={draw=none,anchor=west,align=center,inner sep=0,outer sep=0},
      conv node/.style={rectangle,fill=red!60,draw,font=\sffamily\scriptsize\bfseries,align=center,anchor=west,minimum height=1.5cm},
      blob node/.style={ellipse,fill=gray!20,draw,font=\sffamily\scriptsize\bfseries,align=center,inner sep=0},
      elps node/.style={fill=none,draw=none,font=\sffamily\LARGE\bfseries}]
      
      \node[input node] (lstm_In3) at (0.0,9.0) {\includegraphics[width=2cm]{tan_mug_400_21_data0270.png}};
      \node[input node] (lstm_Rec1) at (0.0,7.225) {\includegraphics[width=2cm]{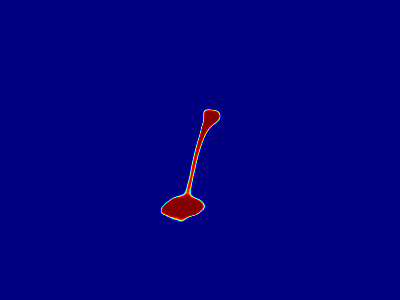}};

      \node[conv node] (lstm_Conv1) at (2.2,8.975) {\rotatebox{270}{\parbox[c]{1.45cm}{\centering Convolution\\{\normalfont\tiny\it 32 $5{\times}5$ kernels}}}};
      \node[conv node] (lstm_Conv2) at (3.1,8.975) {\rotatebox{270}{\parbox[c]{1.45cm}{\centering Convolution\\{\normalfont\tiny\it 32 $5{\times}5$ kernels}}}};
      \node[conv node] (lstm_Conv3) at (4.0,8.975) {\rotatebox{270}{\parbox[c]{1.45cm}{\centering Convolution\\{\normalfont\tiny\it 32 $5{\times}5$ kernels}}}};
      \node[conv node] (lstm_Conv4) at (4.9,8.975) {\rotatebox{270}{\parbox[c]{1.45cm}{\centering Convolution\\{\normalfont\tiny\it 32 $5{\times}5$ kernels}}}};
      \node[conv node] (lstm_Conv5) at (5.8,8.975) {\rotatebox{270}{\parbox[c]{1.7cm}{\centering Convolution\\{\normalfont\tiny\it 32 $17{\times}17$ kernels}}}};
      
      \node[conv node] (lstm_rec_conv1) at (2.2,7.2) {\rotatebox{270}{\parbox[c]{1.45cm}{\centering Convolution\\{\normalfont\tiny\it 20 $5{\times}5$ kernels}}}};
      \node[conv node] (lstm_rec_conv2) at (3.1,7.2) {\rotatebox{270}{\parbox[c]{1.45cm}{\centering Convolution\\{\normalfont\tiny\it 20 $5{\times}5$ kernels}}}};
      \node[conv node] (lstm_rec_conv3) at (4.0,7.2) {\rotatebox{270}{\parbox[c]{1.45cm}{\centering Convolution\\{\normalfont\tiny\it 20 $5{\times}5$ kernels}}}};
      
      \node[conv node] (lstm_lstm1) at (7.0,8.0) [fill=green!60]{\parbox[c]{0.65cm}{\centering LSTM\\{\normalfont\tiny\it 20~$1{\times}1$ kernels \\\vspace{-0.1cm}per~gate}}};
      \node[conv node] (lstm_fc_conv1) at (8.4,8.0) [fill=blue!60]{\rotatebox{270}{\parbox[c]{2.1cm}{\centering $\mathbf{1{\times}1}$ Convolution\\{\normalfont\tiny\it 64 kernels}}}};
      \node[conv node] (lstm_deconv) at (9.3,8.0) [fill=orange!60]{\rotatebox{270}{\parbox[c]{2.0cm}{\centering Deconvolution\\{\normalfont\tiny\it 64 $16{\times}16$ kernels}}}};
      \node[input node] (lstm_Out1) at (10.2,8.0) {\includegraphics[width=2cm]{tan_mug_400_21_pred0270.png}};

      \node[blob node] (lstm_rec_in2) at (6.0, 7.0) {Recurrent\\State};
      \node[blob node] (lstm_rec_in3) at (7.45, 6.7) {Cell\\State};
      
      \node[blob node] (lstm_rec_out2) at (8.7, 9.5) {Recurrent\\State};
      \node[blob node] (lstm_rec_out3) at (7.45, 9.3) {Cell\\State};
      
      \draw (lstm_In3) -- (lstm_Conv1);
      \draw (lstm_Conv1) -- (lstm_Conv2);
      \draw (lstm_Conv2) -- (lstm_Conv3);
      \draw (lstm_Conv3) -- (lstm_Conv4);
      \draw (lstm_Conv4) -- (lstm_Conv5);
      
      \draw (lstm_Rec1) -- (lstm_rec_conv1);
      \draw (lstm_rec_conv1) -- (lstm_rec_conv2);
      \draw (lstm_rec_conv2) -- (lstm_rec_conv3);

      \draw (lstm_Conv5.east) -- (lstm_lstm1.west);
      \draw (lstm_rec_conv3.east) -- (lstm_lstm1.west);
      
      \draw (lstm_lstm1) -- (lstm_fc_conv1);
      \draw (lstm_fc_conv1) -- (lstm_deconv);
      \draw (lstm_deconv) -- (lstm_Out1);
      
      \draw (lstm_rec_in2) -- (lstm_lstm1.west);

      \draw (lstm_rec_in3) -- (lstm_lstm1.south);
      \draw (lstm_lstm1.east) -- (lstm_rec_out2.215);
      \draw (lstm_lstm1.north) -- (lstm_rec_out3);

      \node[conv node] (concat) at (7.5,3.525) [fill=gray!60]{\rotatebox{270}{\parbox[c]{2.0cm}{\centering Concatenation\\{\normalfont\tiny\it channel-wise}}}};
      \node[conv node] (Conv6) at (8.5,3.525) {\rotatebox{270}{\parbox[c]{1.45cm}{\centering Convolution\\{\normalfont\tiny\it 32 $5{\times}5$ kernels}}}};
      \node[conv node] (fc_conv1) at (9.5,3.525) [fill=blue!60]{\rotatebox{270}{\parbox[c]{2.0cm}{\centering Fully Connected\\{\normalfont\tiny\it 256 nodes}}}};
      \node[conv node] (fc_conv2) at (10.5,3.525) [fill=blue!60]{\rotatebox{270}{\parbox[c]{2.0cm}{\centering Fully Connected\\{\normalfont\tiny\it 256 nodes}}}};
      \node[conv node] (fc_conv3) at (11.5,3.525) [fill=blue!60]{\rotatebox{270}{\parbox[c]{2.0cm}{\centering Fully Connected\\{\normalfont\tiny\it 100 nodes}}}};

      \node[output node] (Out1) at (12.5,3.525) {\includegraphics[width=2cm]{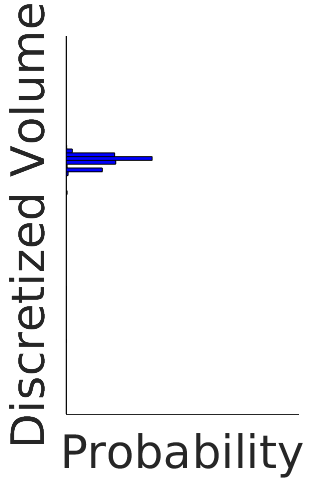}};

      \node[input node] (In3a) at (0.0,4.0) {\includegraphics[width=2cm]{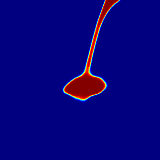}};
      \node[conv node] (Conv1a) at (2.2,3.975) {\rotatebox{270}{\parbox[c]{1.45cm}{\centering Convolution\\{\normalfont\tiny\it 32 $5{\times}5$ kernels}}}};
      \node[conv node] (Conv2a) at (3.1,3.975) {\rotatebox{270}{\parbox[c]{1.45cm}{\centering Convolution\\{\normalfont\tiny\it 32 $5{\times}5$ kernels}}}};

      \node[conv node] (Conv3a) at (4.0,3.975) {\rotatebox{270}{\parbox[c]{1.45cm}{\centering Convolution\\{\normalfont\tiny\it 32 $5{\times}5$ kernels}}}};
      \node[conv node] (Conv4a) at (4.9,3.975) {\rotatebox{270}{\parbox[c]{1.45cm}{\centering Convolution\\{\normalfont\tiny\it 32 $5{\times}5$ kernels}}}};
      \node[conv node] (Conv5a) at (5.8,3.975) {\rotatebox{270}{\parbox[c]{1.7cm}{\centering Convolution\\{\normalfont\tiny\it 32 $17{\times}17$ kernels}}}};
      \draw (In3a) -- (Conv1a);

      \draw (Conv1a) -- (Conv2a);
      \draw (Conv2a) -- (Conv3a);
      \draw (Conv3a) -- (Conv4a);
      \draw (Conv4a) -- (Conv5a);
      \draw (Conv5a) -- (concat);

      \node (elps_node) at (0.22,2.5) [fill=none,draw=none]{\rotatebox{295}{............}};

      \node[input node] (In3b) at (0.5,3.0) {\includegraphics[width=2cm]{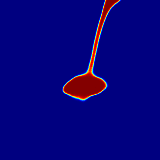}};
      \node[conv node] (Conv1b) at (2.7,3.075) {\rotatebox{270}{\parbox[c]{1.45cm}{\centering Convolution\\{\normalfont\tiny\it 32 $5{\times}5$ kernels}}}};
      \node[conv node] (Conv2b) at (3.6,3.075) {\rotatebox{270}{\parbox[c]{1.45cm}{\centering Convolution\\{\normalfont\tiny\it 32 $5{\times}5$ kernels}}}};
      \node[conv node] (Conv3b) at (4.5,3.075) {\rotatebox{270}{\parbox[c]{1.45cm}{\centering Convolution\\{\normalfont\tiny\it 32 $5{\times}5$ kernels}}}};

      \node[conv node] (Conv4b) at (5.4,3.075) {\rotatebox{270}{\parbox[c]{1.45cm}{\centering Convolution\\{\normalfont\tiny\it 32 $5{\times}5$ kernels}}}};
      \node[conv node] (Conv5b) at (6.3,3.075) {\rotatebox{270}{\parbox[c]{1.7cm}{\centering Convolution\\{\normalfont\tiny\it 32 $17{\times}17$ kernels}}}};
      \draw (In3b) -- (Conv1b);
      \draw (Conv1b) -- (Conv2b);
      \draw (Conv2b) -- (Conv3b);

      \draw (Conv3b) -- (Conv4b);
      \draw (Conv4b) -- (Conv5b);
      \draw (Conv5b) -- (concat);

      \draw (concat) -- (Conv6);
      \draw (Conv6) -- (fc_conv1);
      \draw (fc_conv1) -- (fc_conv2);
      \draw (fc_conv2) -- (fc_conv3);
      \draw (fc_conv3) -- (Out1);
      
      
      \node[conv node] (crop) at (5.0,5.7) [fill=magenta!60,minimum height=0.7cm]{\parbox[c]{2.0cm}{\centering Crop\\{\normalfont\tiny\it $160{\times}160$}}};
      \draw[line width=0.1cm] (lstm_Out1.south) to [out=270,in=0] (crop.east);
      \draw[line width=0.1cm] (crop.west) to [out=180,in=90] (In3a.north);
      
      \node[conv node] (hmm) at (4.0,0.0) [fill=cyan!60,minimum height=0.7cm]{\parbox[c]{2.0cm}{\centering HMM}};
      \node[conv node] (pid) at (8.0,0.0) [fill=violet!60,minimum height=0.7cm]{\parbox[c]{2.0cm}{\centering PID \\Controller}};
      \node[draw=none] (control) at (12.0,0.0) {\parbox[c]{1.2cm}{\centering Robot\\Control\\Signal}};
      \draw[line width=0.1cm] (Out1.south) to [out=270,in=90] (hmm.north);
      \draw[line width=0.1cm] (hmm.east) -- (pid.west);
      \draw[line width=0.1cm] (pid.east) -- (control.west);
      \draw[line width=0.1cm] (hmm.east) -- (7.0,0.0) -- (7.0,-1.0) -- (3.0,-1.0) -- (3.0,0.0) -- (hmm.west);
      \node[draw=none] at (4.7,0.7) {{\bf\Large$z_t$}}; 
      \node[draw=none] at (6.6,0.4) {{\bf\Large$v_t$}};
      
\end{tikzpicture}
\vspace{-0.5cm}

\caption{The entire robot control system using the recurrent neural network for detections and the multi-frame network for volume estimation. The recurrent detection network (top) takes both the color image and its own detections from the previous time step and produces a liquid detection heatmap. The multi-frame network (center) takes a sequence of detections cropped around the target container and outputs a distribution over volumes in the container. The output of this network is fed into a HMM, which estimates the volume of the container. This is passed into a PID controller, which computes the robot's control signal.}
\label{fig:model}
\vspace{-0.75cm}
\end{center}

\end{figure*}

Given color images, we train a convolutional neural network (CNN) to label each pixel as {\it liquid} or {\it not-liquid} (we use the thermal camera to acquire the ground truth labeling). 
The network is fully-convolutional, that is, all the learned layers are convolution layers with no fully-connected layers. 
The output of the network is a heatmap over the image, with real values in the range $[0,1]$, where higher values indicate a higher likelihood of liquid.
In \cite{schenckc2016b} we tested 3 network structures and found that a recurrent network utilizing a long short-term memory (LSTM) layer \cite{hochreiter1997} performs the best of the 3. 
Here we use the LSTM-CNN from that paper, which is shown in the top row of Figure \ref{fig:model}. 
We refer the reader to \cite{schenckc2016b} for more details.

\subsection{Volume Estimation During Pouring Sequences}

We propose two different methods for estimating the volume of liquid in a target container. 
The first is a model-based method, which assumes we have access to a 3D model of the target container and infers the height of the liquid based on the camera pose and binary pixel labels. 
The second is a model-free method that trains a neural network to regress to the volume of liquid in the target container given labeled pixels.

\subsubsection{Filtering using a HMM}
\label{sec:hmm}

Before describing our volume estimation methods, we first describe our filtering method, which will make our notation in the following sections clearer.
Because of the temporal nature of the task, we utilize a hidden Markov model (HMM) to filter the volume estimates over time.
Let $t$ be the current timestep, $v_t$ be the volume of liquid in the target container at time $t$ (the hidden state in the HMM), and let $z_t$ be the observation at time $t$ (described in detail in the following sections).
To compute the probability distribution over $v_t$ we can apply Bayes rule.
HMMs make the Markovian assumption, that is, $v_t$ is conditionally independent of all prior observations and states given $v_{t-1}$ and $z_t$ is conditionally independent of all prior observations and states given $v_t$.
Thus we can write the posterior as
\[ P(v_t | z_t, v_{t-1} ) \propto P(z_t | v_t ) P(v_t | v_{t-1} ). \]
For this paper, we represent the distribution over $v$ as a histogram over a fixed range, and so the transition probability $P(v_t | v_{t-1} )$ is a summation over the bins in the histogram
\begin{align}
\label{eqn:prior}
P(v_t | v_{t-1} )  = \displaystyle\sum_i P( v_t | v_{t-1} = i) P( v_{t-1} = i ).
\end{align}
The transition probability $P( v_t | v_{t-1} )$ is inferred from the training data. 

The following two sections describe how we compute the observation probability $p(z_t | v_t)$, which varies for the model-based and model-free methods. 
During task execution, we compute the volume of liquid at a given timestep $t$ by taking the median over the posterior distribution on $v_t$\footnotemark.

\subsubsection{Model-Based Volume Estimation}
\label{sec:model-based}

Our model-based method for estimating the volume of liquid in a target container assumes we have a 3D model of the container and that we can use the pointcloud from our RGBD sensor to find its pose in the scene.
The observation $z_t$ for this method is the set of pixel-wise liquid labels for the image at time $t$, computed as described in section \ref{sec:classification}.
Intuitively, to compute the observation probabilities, this method compares the actual observation $z_t$ to what the robot would expect to see if the volume of liquid in the container were $v_t$.

\footnotetext{We also evaluated using other methods such as the expectation or maximum likelihood, but we empirically determined that median produced more stable and less skewed estimates, although all the methods only had minor differences.}

\begin{figure}
\begin{center}
\vspace{0.5cm}
\scalebox{0.5}{
\begin{tikzpicture}[>=stealth]
    
    \draw[draw=none,fill=gray!30!white,rotate=-45] (-3cm,5cm) -- (3cm,2.0cm) -- (3cm,2.2cm) -- (-0.1cm,4.05cm) -- (-3cm,5cm);
    \draw[draw=none,fill=gray!30!white,rotate=-45] (-3cm,5cm) -- (3cm,8cm) -- (3cm,6.45cm) -- (-3cm,5cm);
    \draw[line width=1.0mm,dashed,draw=black!15!orange,fill=black!15!orange,->] (1.4cm,5.65cm) -- (5.7cm,2.9cm);
    \draw[line width=1.0mm,dashed,draw=black!15!orange,fill=black!15!orange,->] (1.4cm,5.65cm) -- (5.53cm,2.2cm);
    \draw[line width=1.0mm,dashed,draw=black!15!blue,fill=black!15!blue,->] (1.4cm,5.65cm) -- (5.0cm,2.0cm);
    \draw[line width=1.0mm,dashed,draw=black!15!blue,fill=black!15!blue,->] (1.4cm,5.65cm) -- (4.0cm,2.0cm);
    \draw[fill=black,draw=none,rotate=-45,rounded corners] (-4.5cm,4.5cm) rectangle (-2.5cm,5.5cm);
    \draw[draw=none,fill=black,rotate=-45] (-3cm,5cm) -- (-2cm,5.5cm) -- (-2cm,4.5cm) -- (-3cm,5cm);

    \draw[draw=none,fill=blue,rounded corners] (3.0cm,2cm) -- (3.5cm,0cm) -- (5cm,0cm) -- (5.5cm,2cm) -- (3.0cm,2cm); 
    \draw[line width=1.0mm,rounded corners,draw=black!15!orange] (2.75cm,3cm) -- (3.5cm,0cm) -- (5cm,0cm) -- (5.75cm,3cm); 
    
    \draw[line width=1.0mm,draw=black!30!black,dashed] (2cm,2cm) -- (6cm,2cm);  
    \node[draw=none] at (1.5cm,2cm) {{\Huge $h_t$}}; 
    
\end{tikzpicture}
}
\vspace{-0.5cm}
\end{center}
\caption{Diagram of the camera looking into the target container. The blue shows where the liquid is expected to be, given a volume and the corresponding fill height, $h_t$. The blue lines show pixels expected to be classified as {\it liquid} and the orange lines show pixels that are expected to be classified as {\it not-liquid}. The gray shows pixels outside the container and not used by our algorithm.}
\label{fig:therm_layout}
\vspace{-0.75cm}
\end{figure}
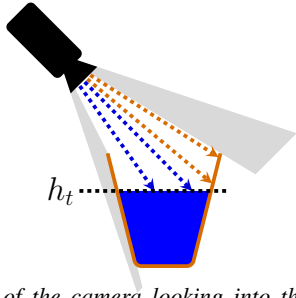

More formally, we compute $P( z_t | v_t = i )$ as follows.
First, we assume conditional independence between every pixel ${p_t^j \in z_t}$.
Thus the observation probability becomes
\[ P( z_t | v_t = i ) = \displaystyle\prod_j P(p_t^j | v_t = i). \]
For this product, we only consider the set of pixels that view the inside of the container, i.e., the set of pixels that could potentially be labelled liquid.
An example of this is shown in Figure \ref{fig:therm_layout}.
The dashed lines show the pixels whose rays intersect the interior of the container, whereas the gray areas represent the pixels whose rays do not.
Since the pixels in the gray area can not see liquid in the container, they have no effect on the observation probability and thus are not considered.

To compute $P( p_t^j | v_t = i )$, we use the 3D mesh of the container and fill it with $v_t = i$ volume of liquid.
We then project that liquid back into the camera to get the expected pixel label $\widehat{p}_t^j$.
The observation probability is then
\begin{align*}
P( p_t^j | v_t = i ) = & P(p_t^j | \widehat{p}_t^j = \text{{\it liquid}}) P(\widehat{p}_t^j = \text{{\it liquid}} | v_t = i) \; + \\
P(p_t^j | \widehat{p}_t^j &= \text{{\it not-liquid}}) P(\widehat{p}_t^j = \text{{\it not-liquid}} | v_t = i).
\end{align*}
To compute $P(\widehat{p}_t^j | v_t = i)$, we assume that the liquid is resting level in the container\footnote{While not strictly true throughout the entire duration of a pour, this assumption still allows for a good measurement.}.
At rest, the surface of the liquid will be parallel to the ground, and so we find the height $h_t$ of the surface.
We place a plane parallel to the ground at $h_t$ and check whether the ray from pixel $p_t^j$ intersects that plane prior to intersecting the 3D mesh of the container.
We set $P(\widehat{p}_t^j = \text{{\it liquid}} | v_t = i)$ to be 1 if the ray intersects the plane first (and 0 otherwise), and we set $P(\widehat{p}_t^j = \text{{\it not-liquid}} | v_t = i)$ to be 1 if the ray intersects the mesh first (and 0 otherwise).
Figure \ref{fig:therm_layout} shows an example.
The blue dashed lines show pixels that intersect the plane, whereas the orange dashed lines show pixels that intersect the mesh before intersecting the plane.
To compute $P(p_t^j | \widehat{p}_t^j)$, we use the following table:
\begin{center}
\begin{tabular}{ r r || c c }
   &  & \multicolumn{2}{c}{{\bf $p_t^j$}} \\
   &  & {\it Liquid} & {\it Not-liquid} \\
     \hline\hline
 \multirow{2}{*}{{\bf $\widehat{p}_t^j$}} & {\it Liquid} & 90\% & 10\% \\
  & {\it Not-liquid} & 20\% & 80\% \\
\end{tabular}
\end{center}

To compute the height of the surface of the liquid $h_t$, we use binary search in combination with the signed tetrahedron volume method \cite{zhang2001}.
That is, given a height $h_t$, we can compute the volume of the interior of the container below that height using its 3D mesh by applying the method described in \cite{zhang2001}.
This volume under the plane corresponds to the volume of liquid resting in the container.
We then perform binary search over the height to find the $h_t$ that corresponds to $v_t = i$.
Note that because the distribution over $v_t$ is a histogram with a fixed set of bins, we can precompute the height for each value of $v_t$ for each 3D mesh.


\subsubsection{Model-Free Volume Estimation}

Our model-free method replaces the object pose inference of the model-based method with a neural network. 
The neural network takes in pixel labels and produces a volume estimate. 
We use only the output of the detection network described in section \ref{sec:classification} for the pixel labels, so we directly feed the heatmap over the pixels into the volume estimation network. 
We also evaluate adding as inputs either the color or depth images, which we append channel-wise to the pixel labels before feeding into the network. 
We crop the input to the network around the target container. 

Formally, the network computes the function $f(z_t) = \tilde{v}_t$, where $z_t$ is the observation and $\tilde{v}_t$ is an estimate of the volume in the container.
The estimate $\tilde{v}_t$ is used as the observation in the HMM, i.e., we use $P(\tilde{v}_t | v_t)$ in place of $P(z_t | v_t)$ since $\tilde{v}_t$ is a function of the observation $z_t$.
The output of the network is a distribution over $\tilde{v}_t$.
To compute $P(\tilde{v}_t | v_t)$, we sum over all values for $\tilde{v}_t$
\[ P(\tilde{v}_t | v_t) = \displaystyle\sum_k P(v_t | \tilde{v}_t = k) P(\tilde{v}_t = k) \]
where $P(\tilde{v}_t = k)$ is computed by the network.
The conditional probability $P(v_t | \tilde{v}_t = k)$ is inferred from the training data.

We evaluated three different network architectures: a single-frame CNN, a multi-frame CNN, and a recurrent LSTM CNN. 
We use the Caffe deep learning framework \cite{jia2014} to implement our networks

{\it Single-Frame CNN:}
The single-frame network is a standard CNN that takes as input a single image.
It then passes the image through 5 convolution layers, each of which is followed by a max pooling and rectified linear layer.
Every layer has a stride of 1 except for the first 3 max pooling layers, which have a stride of 2. 
It passes the result through 3 fully connected layers, each followed by a rectified linear layer. 
These last 3 layers are also followed by dropout layers during training, with a drop rate of 10\%.
The single-frame network (CNN) is similar to the multi-frame network shown in the center row of Figure \ref{fig:model}, with the exception that it only takes a single frame and does not have the concatenation layer or the convolution layer immediately following it.

{\it Multi-Frame CNN:}
The multi-frame network (MF-CNN) is shown in the center row of Figure \ref{fig:model}. 
It takes as input a set of temporally sequential images. 
Each image is passed independently through the first 5 layers of the network, which are identical to the first 5 convolutional layers in the single-frame network.
Next, the result of each image is concatenated channel-wise and passed through another convolution layer (which is also followed by max pooling and rectified linear layers).
This is then fed into 3 fully connected layers, which are identical to the last 3 layers of the single-frame CNN.

{\it Recurrent LSTM CNN:}
The LSTM-CNN is identical to the single-frame network, with the exception that we replace the first fully connected layer with the LSTM layer.
In addition to the output of the convolution layers, the LSTM layer also takes as input the recurrent state from the previous timestep, as well as the cell state from the previous timestep.
Each gate in the LSTM layer is a 256 node fully connected layer.
Please refer to Figure 1 of \cite{greff2015} for a detailed layout of the LSTM layer.

\subsection{Robot Controller}
\label{sec:robot_controller}

For this paper, we want to investigate whether, given good real-time feedback, pouring can be performed with a simple controller.
We place a table in front of the robot, and on the table we place the target container. 
We fix the source container in the robot's right gripper and pre-fill it with a specific amount of water not given to the robot. 
We also fix the robot's arm such that the source container is above and slightly to the side of the target container.

To pour, the robot controls the angle of its wrist joint, thus directly controlling the angle of the source container. 
We use a modified PID controller to execute the pour. 
The robot first tilts the container to a pre-specified angle (we use 75 degrees from vertical), then begins running the PID controller, using the difference between the target volume and the current volume in the target container as its error signal, which it uses to set the angular velocity of its wrist joint.
Since pouring is a non-reversible task (liquid cannot return to the source once it has left), the integral term does nothing except push the robot to pour faster, so we set its gain to 0\footnote{We refer to the controller as a PID controller for easier comprehension by the reader, but it is technically a PD controller.}. We set the proportional and derivative gains to $\frac{0.01\pi}{180}$ and $\frac{0.2\pi}{180}$ respectively, which we empirically determined to perform well for the pouring task.
Once the target volume has been reached, the robot stops the PID controller and rotates the source container until it is vertical once again.

\vspace{-0.2cm}
\subsection{Implementation Details}
\vspace{-0.0cm}

\subsubsection{Finding the Container in the Scene}

Both our model-based and model-free methods require finding the target container on the table in front of the robot (though only the model-based needs a 3D model).
To find the container, we use the robot's RGBD camera to capture a pointcloud of the scene in front of the robot and then utilize functions in the PointCloud Library (PCL) \cite{rusu2011} to find the plane of the table and cluster the points on top of it.
To acquire the pose for the model-based method, we use iterative closest points to find the 3D pose of the model in the scene.
Next we use this pose to label each pixel in the image as either {\it inner} (inside of the container), {\it outer} (outside of the container), or {\it neither}.

\subsubsection{Generating Ground Truth Pixel Labels}

We use a thermal camera in combination with water heated to approximately 93\degree Celsius to get the ground truth pixel labels for the liquid.
To register the thermal image to the color image, we use a paper checkerboard pattern attached to a $61{\times}61$ centimeter metal aluminum sheet.
We then direct a small, bright spotlight at the pattern, causing a heat differential between the white and black squares, which is visible as a checkerboard pattern in the thermal image. 
We use OpenCV's built-in function for finding corners of a checkerboard to find correspondence points and compute an affine transformation\footnote{While there has been prior work on performing full registration between thermal and color images \cite{pinggera2012}, because the depth of the pour is fixed, we opted for this simpler approach. We cannot use the depth sensor because water does not appear on the depth image.}.
We use an adaptive threshold based on the average temperature of the pixels associated with the target container (which includes the pixels for the liquid in the container).  
The result of this is a binary image with each pixel classified as either {\it liquid} or {\it not-liquid}.
Figure \ref{fig:therm_examples} shows a color image, its corresponding thermal image transformed to the color pixel space, and a simple temperature threshold of the thermal image.
Note that the thermal camera provides quite reliable pixel labels for liquid detection with minimal false positives

\subsubsection{Acquiring Ground Truth Volume Estimates for Training and Evaluation}
\label{sec:get_gt}

In order to train our networks in the previous section, and to evaluate both our model-based and model-free methods, we need a baseline ground truth volume estimation.
To generate this baseline, we utilize the thermal camera in combination with the model-based method described in section \ref{sec:model-based}.
However, since this analysis can be done {\it a posteriori} and does not need to be real-time, we can use the benefit of hindsight to improve our estimates, i.e., future observations can improve the current state estimate.
While we acknowledge that this method does not guarantee perfect volume estimates, the combined accuracy of the thermal camera and after-the-fact processing yield robust estimates suitable for training and evaluation.

To compute this baseline we replace the forward method for HMM inference described in section \ref{sec:hmm} with Viterbi decoding \cite{blunsom2004}.
We replace the summation in equation \ref{eqn:prior} in the computation of the prior $P(v_t | v_{t-1})$ with a $max$ to compute the probability of each sequence.
We use a corresponding $argmax$ to compute the previous state from the current state, starting at the last time step and working backwards.
At the last time step, we start with the most probable state.
Thus using this method we can generate a reliable ground truth estimate of the volume of liquid in the target container over the duration of a pouring sequence to use for training our learning algorithms and evaluating our methodology.
 

\vspace{-0.1cm}
\section{Experiments \& Results}

\subsection{Robotic Platform}

All of our experiments were performed on our Rethink Robotics Baxter Research Robot, shown in Figure \ref{fig:robot_setup}.
It is equipped with two 7-dof arms, each with an electric parallel gripper.
For the experiments in this paper, we use exclusively the right arm.
The robot has an Asus Xtion Pro mounted on its upper-torso, directly below its screen, which includes both an RGB color camera and a depth sensor, each of which produce $640{\times}480$ images at 30Hz.
Mounted on the robot immediately above the Xtion sensor is an Infrared Cameras Inc. 8640P Thermal Imaging Camera, which reads the temperature of the image at each pixel and outputs a $640{\times}512$ image at 30Hz.

\subsection{Experimental Setup}

For all experiments, the robot poured from the cup shown in its gripper in Figure \ref{fig:robot_setup}. 
We used three target containers, also shown in Figure \ref{fig:robot_setup}. 
We collected a dataset of pours using this setup in order to both train and evaluate our methodologies. 
We collected a total of 279 pouring sequences, in which the robot attempted to pour 250ml of water into the target using the thermal camera with the model-based method, with the initial amount in the cup varied between 300ml, 350ml, and 400ml.
Each sequence lasted exactly 25 seconds and was recorded on both the thermal and RGBD cameras at 30Hz.
We randomly divided the data 75\%-25\% into train and evaluation sets.
After the data was collected, we used the thermal images to generate ground truth pixel labels as well as we used the Viterbi decoding method described in section \ref{sec:get_gt} to generate ground truth volume estimates, which we compare against for the remainder of this section.

\subsection{Validating Thermographic Ground Truth Methodology}

\begin{figure}
    \centering
    \setlength{\unitlength}{1.0cm}
    \begin{subfigure}{4.0cm}
        \begin{picture}(4.0,3.5)
            \put(0.0,0.0){\includegraphics[width=4.0cm]{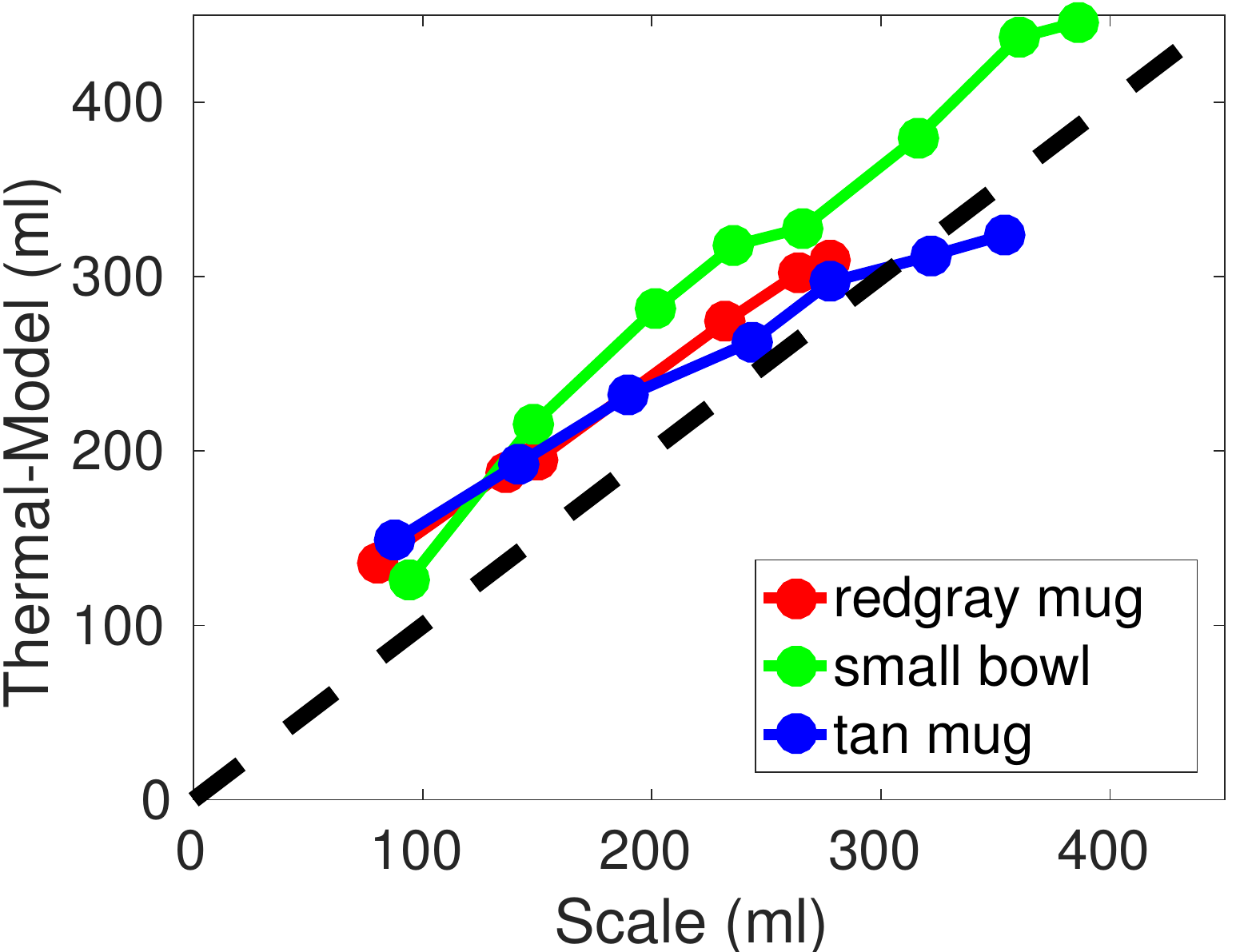}}
        \end{picture}
        \caption{Scale vs. Thermal}
        \label{fig:thermal_verification}
    \end{subfigure}\hspace{0.5cm}%
    \begin{subfigure}{4.0cm}
        \begin{picture}(4.0,3.5)
            \put(0.0,0.0){\includegraphics[width=4.0cm]{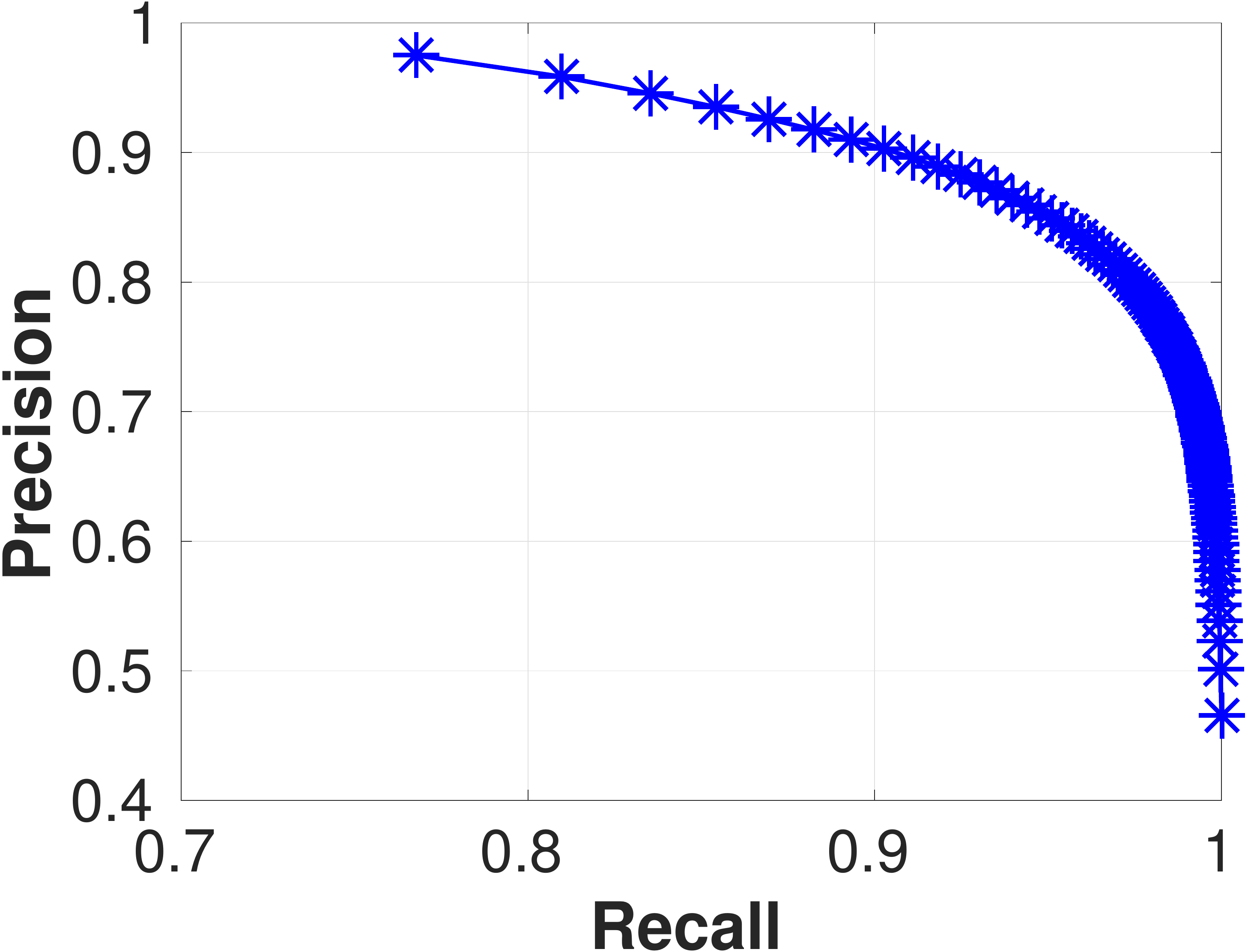}}
        \end{picture}
        \caption{Detection Network}
        \label{fig:detection_verification}
    \end{subfigure}
    \caption{The left plot shows the scale reading compared to the thermal camera with the model based method for each of the target containers. The right plot shows the pixel-wise precision and recall curves for the liquid detection network.}
    \vspace{-0.75cm}
\end{figure}

Before we can evaluate our methodologies, we must first verify that our method for generating ground truth volume estimates is accurate.
We can compare a static volume measurement with a scale to static estimates from the thermal camera combined with the model-based method to gauge the accuracy of our method.
Figure \ref{fig:thermal_verification} shows a comparison between measurements from a scale (x-axis) and the corresponding measurement from the thermal camera using the model-based method (y-axis) for each of the three target containers.
The black dashed line shows a 1:1 correspondence for reference.
From the figure it is clear that the model-based method overestimates the volume for each container.
In order to make our baseline as accurate as possible, we fit a linear model for each container and use that to calibrate the baseline ground truth estimates described in section \ref{sec:get_gt}.

Note that we only use this calibration for computing the ground truth baseline,
and not when computing estimates for the model-based methodology.  This is done
on purpose, since a robot would not be able to pre-calibrate its estimates for
every object in a household setting.  While it might be reasonable to assume
that a robot can acquire a 3D model for each target container via existing
object databases, we believe that performing a pre-calibration using a scale and
multiple pouring experiments for each object would be overly demanding.

\vspace{-0.2cm}
\subsection{Evaluating the Detection Network}

Next we must verify that the neural network we trained to labels pixels as {\it liquid} or {\it not-liquid} from color images is accurate enough to utilize for volume estimation.
While our prior work \cite{schenckc2016b} showed that neural networks can label
liquid pixels reasonably well on data generated by a realistic liquid simulator,
these networks performed poorly on real imagery. Here, we take advantage of our
thermal camera system to train on real data.
We trained the recurrent LSTM CNN using the mini-batch gradient descent method Adam \cite{kingma2014} with a learning rate of 0.0001 and default momentum values, for 61,000 iterations.
We unrolled the recurrent network during training for 32 frames and used a batch size of 5.
We scaled the input images to $400 \times 300$ resolution.
The error signal was computed using softmax loss.
As in \cite{schenckc2016b}, we found the best results are achieved when we first pre-train the network on crops of liquid in the images, and then train on full images.

Figure \ref{fig:detection_verification} shows the performance of the detection network, and the image in Figure \ref{fig:detection_example} shows an example of the output of the network.
These results clearly show that our detection network is able to classify pixels with high precision and recall.
This suggests that the network will work well for estimating the volume of liquid in the target container.
We should note, however, that due to the relatively small size of the training set, this detection network will work well only for the tasks described in this paper and will not generalize to other environments or tasks.

\vspace{-0.2cm}
\subsection{Comparing Methods for Volume Estimation}

\begin{figure}
    \centering
    \setlength{\unitlength}{1.0cm}
    \begin{subfigure}{2.4cm}
        \begin{picture}(2.0,3.5)
            \put(0.0,0.1){\includegraphics[width=2.0cm]{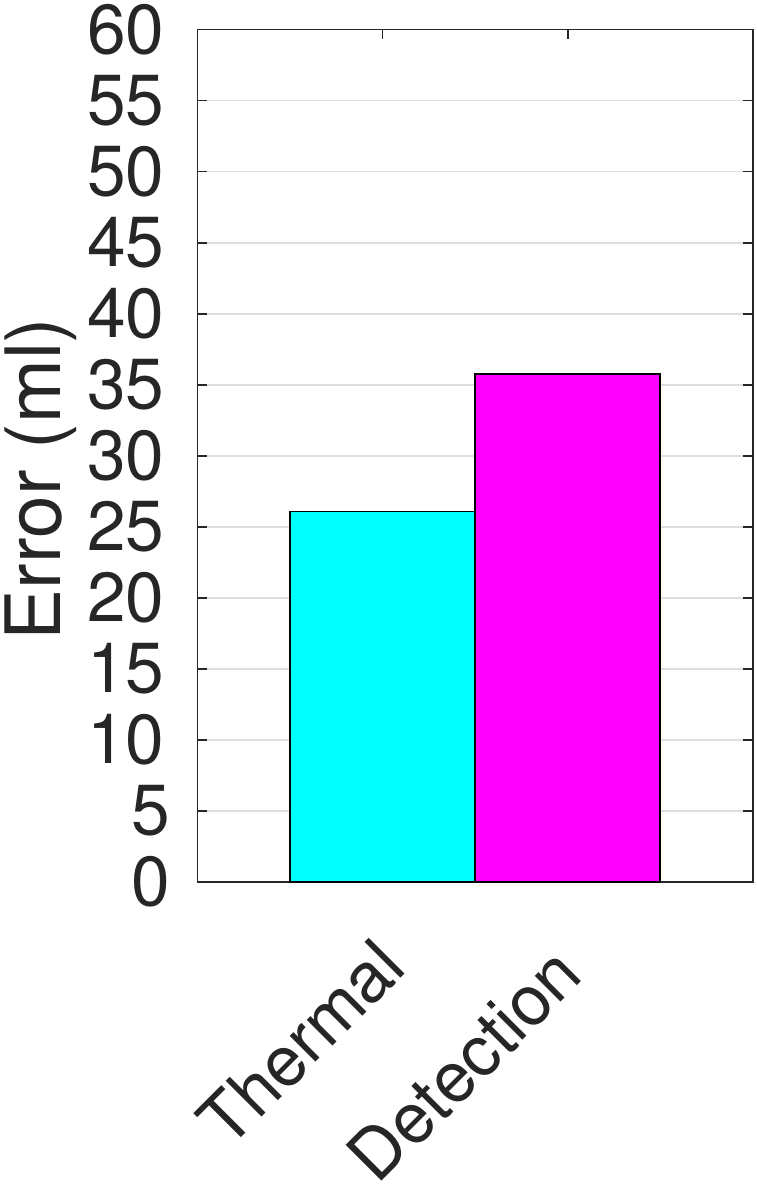}}
        \end{picture}
        \caption{Model-Based}
        \label{fig:model_based}
    \end{subfigure}\hspace{0.1cm}%
    \begin{subfigure}{6.0cm}
        \begin{picture}(6.0,3.5)
            \put(0.0,0.0){\includegraphics[width=6.0cm]{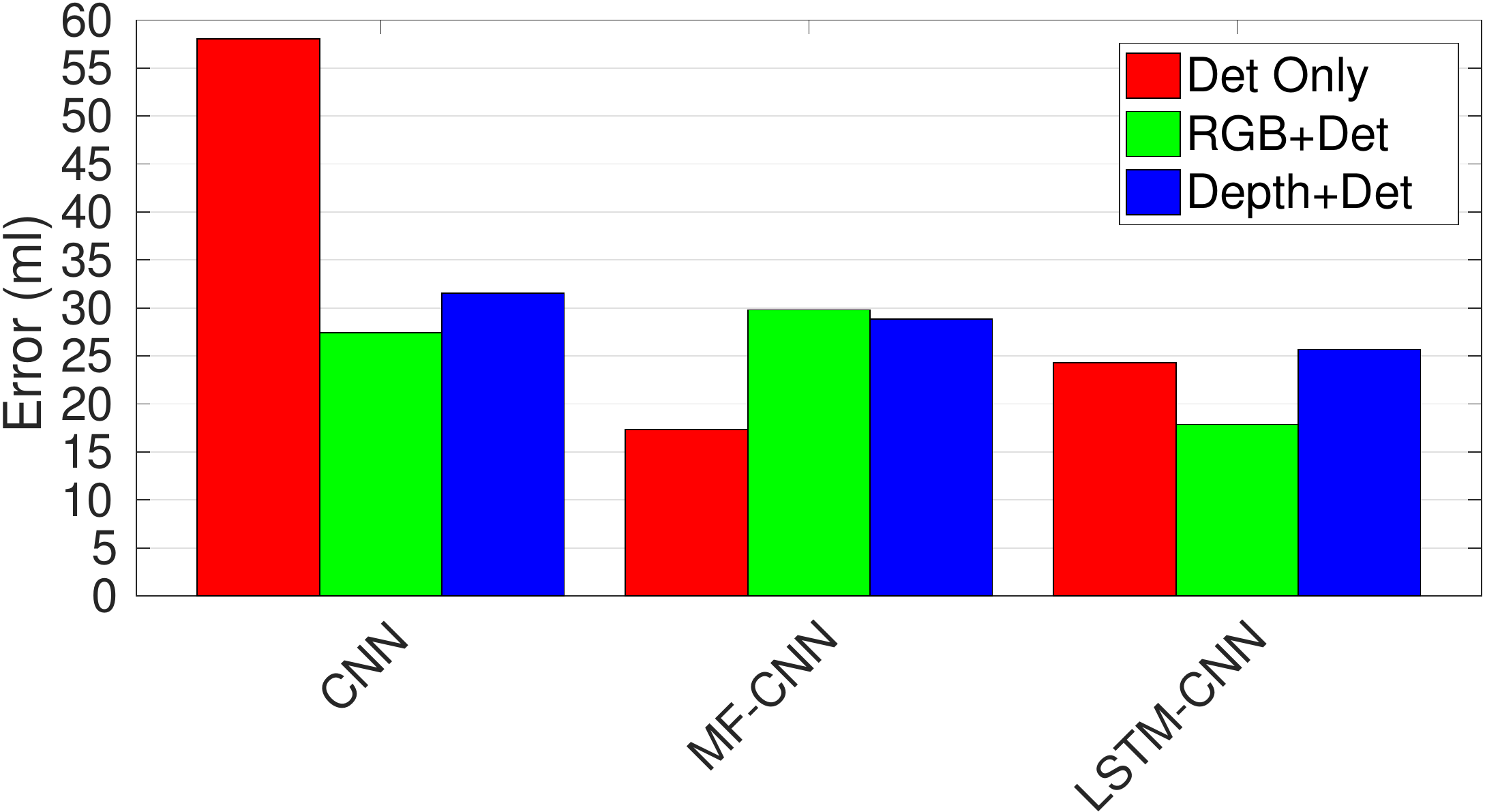}}
        \end{picture}
        \caption{Model-Free}
        \label{fig:model_free}
    \end{subfigure}
    \caption{Root mean squared error in milliliters of each of the methods for volume estimation. The left plot shows the error for the model-based method using either the thermal image or the output of the detection network as pixel labels. The right plot shows the error for the model-free methods when the networks take as input only the liquid detections (red), the liquid detections plus the color image (green), and the liquid detections plus the depth image (blue).}
    \label{fig:aggregate_all}
    \vspace{-0.75cm}
\end{figure}

\begin{figure}
    \centering
    \setlength{\unitlength}{1.0cm}
    \begin{subfigure}{3.0cm}
        \begin{picture}(3.0,2.7)
            \put(0.0,0.0){\includegraphics[width=3.0cm]{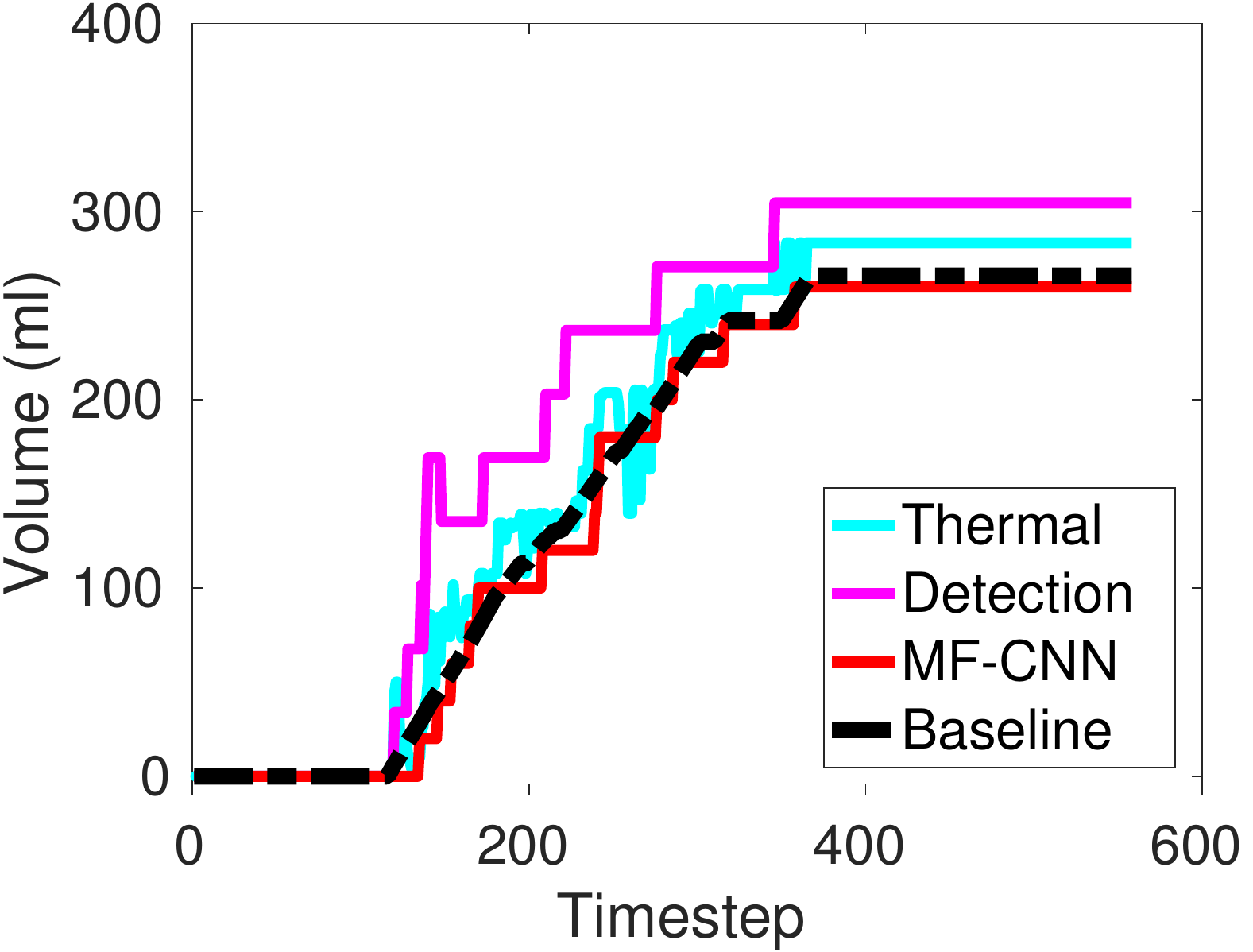}}
        \end{picture}
        \caption{Small bowl}
        \label{fig:ex_small_bowl}
    \end{subfigure}%
    \begin{subfigure}{2.7cm}
        \begin{picture}(2.7,2.7)
            \put(0.0,0.0){\includegraphics[width=2.7cm]{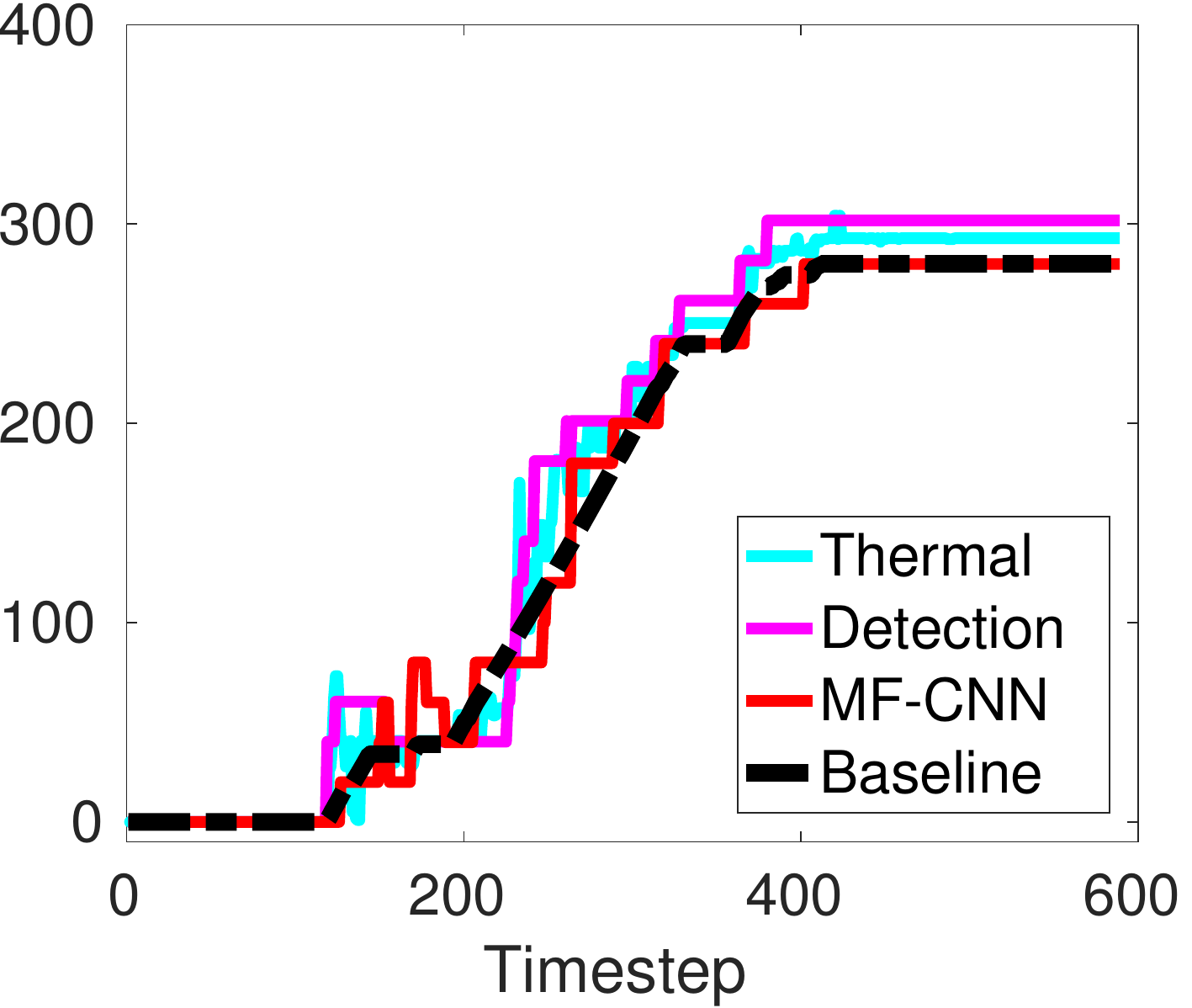}}
        \end{picture}
        \caption{Tan mug}
        \label{fig:ex_tan_mug}
    \end{subfigure}%
    \begin{subfigure}{2.7cm}
        \begin{picture}(2.7,2.7)
            \put(0.0,0.0){\includegraphics[width=2.7cm]{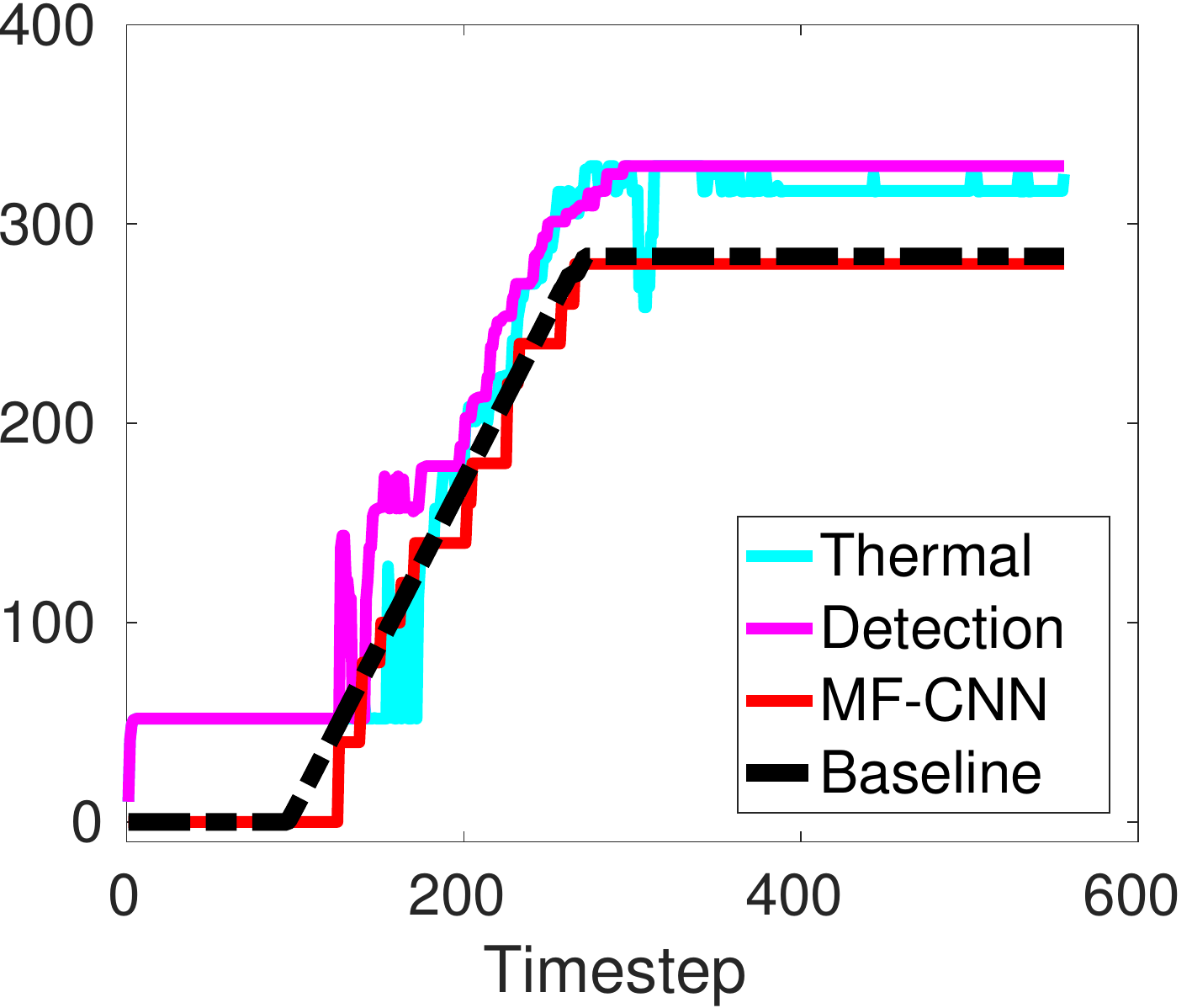}}
        \end{picture}
        \caption{Redgray mug}
        \label{fig:ex_redgray_mug}
    \end{subfigure}
    \caption{The volume estimates for the two model-based methods and the multi-frame detection only model-free network. The black dashed line is the baseline ground truth. We randomly selected one sequence for each target container from our test set to display here. Best viewed in color.}
    \label{fig:exs}
    \vspace{-0.75cm}
\end{figure}

For our model-free methodology, every network was trained using the mini-batch gradient descent method Adam \cite{kingma2014} with a learning rate of 0.0001 and default momentum values. 
Each network was trained for 61,000 iterations, at which point performance tended to plateau. 
All single-frame networks were trained using a batch size of 32; all multi-frame networks with a window of 32 and batch size of 5; and all LSTM networks with a batch size of 5 and unrolled for 32 frames during training. 
The input to each network was a $160 \times 160$ resolution crop of either the liquid detections only, the color image and detections appended channel-wise, or the depth image and detections appended channel-wise.
We discretize the output to 100 values for the range of 0 to 400ml (none of our experiments use volumes greater than 400ml) and train the network to classify the volume.
The error signal was computed using the softmax with loss layer built into Caffe \cite{jia2014}.
In our data we noticed that approximately $\frac{2}{3}$ of the time during each pouring sequence was spent either before or after pouring had occurred, with little change in the volume.
We found that the best results could be achieved by first pre-training each network on data from the middle of each sequence during which the volume was actively changing, and then training on data sampled from the entire sequence.
We discretize $v_t$ and the output of the network into 20 values\footnote{While this may seem rather coarse, we found it works well in practice.}.

Figure \ref{fig:aggregate_all} shows the root mean squared error in milliliters on the testing data for each method with respect to our baseline ground truth comparison described in section \ref{sec:get_gt}.
It should be noted that although both our baseline ground truth estimate and the thermal estimate in Figure \ref{fig:model_based} are derived from the same data, the difference between the two can be largely attributed to the fact that the baseline method is able to look backwards in time and adjust its estimates, whereas the thermal model-based method can only look forward (which is necessary for control).
For example, in the initial frames of a pour, as the water leaves the source container, it can splash against the side of the target container, causing the forward thermal estimate to incorrectly estimate a spike in the volume of liquid, whereas the baseline method can smooth this spike by propagating backwards in time.

While the error for both model-based methods are relatively small, it is clear that some of the model-free methods are actually better able to estimate the volume of liquid in the target container.
Surprisingly, the best performing model-free estimation network is the multi-frame network that takes as input only the pixel-wise liquid detections from the detection network.
The networks trained on detections only are the only networks that receive no shape information about the target container (both the depth and color images contain some information about shape), so intuitively, it would be expected that they would be unable to estimate the volume of more than a single container, and thus perform more poorly than the other networks.
However, a lot of the temporal and perceptual information used by our methodology is already provided in the pixel-wise liquid detections, thus temporal information in addition to either color or depth images are not as beneficial to the networks.

We can verify that this is indeed the case by looking at the volume estimates on randomly selected pouring sequences from the test set, one for each target container.
Figure \ref{fig:exs} shows the volume estimates for the two model-based methods and the multi-frame detection only method as compared to the baseline.
It is clear from the plots that the multi-frame network is better able to match the baseline ground truth than either of the model-based methods.
Not only does the multi-frame network outperform the model-based methods, but unlike them, it does not require either an expensive thermal camera or a model of the target container.
For these reasons, we utilize this method in the next section for carrying out actual pouring experiments with closed-loop visual feedback.

\subsection{Pouring with Raw Visual Feedback}

\begin{figure}
    \centering
    \setlength{\unitlength}{1.0cm}
    \begin{picture}(7.0,5.6)
        \put(0.0,0.0){\includegraphics[width=7.0cm]{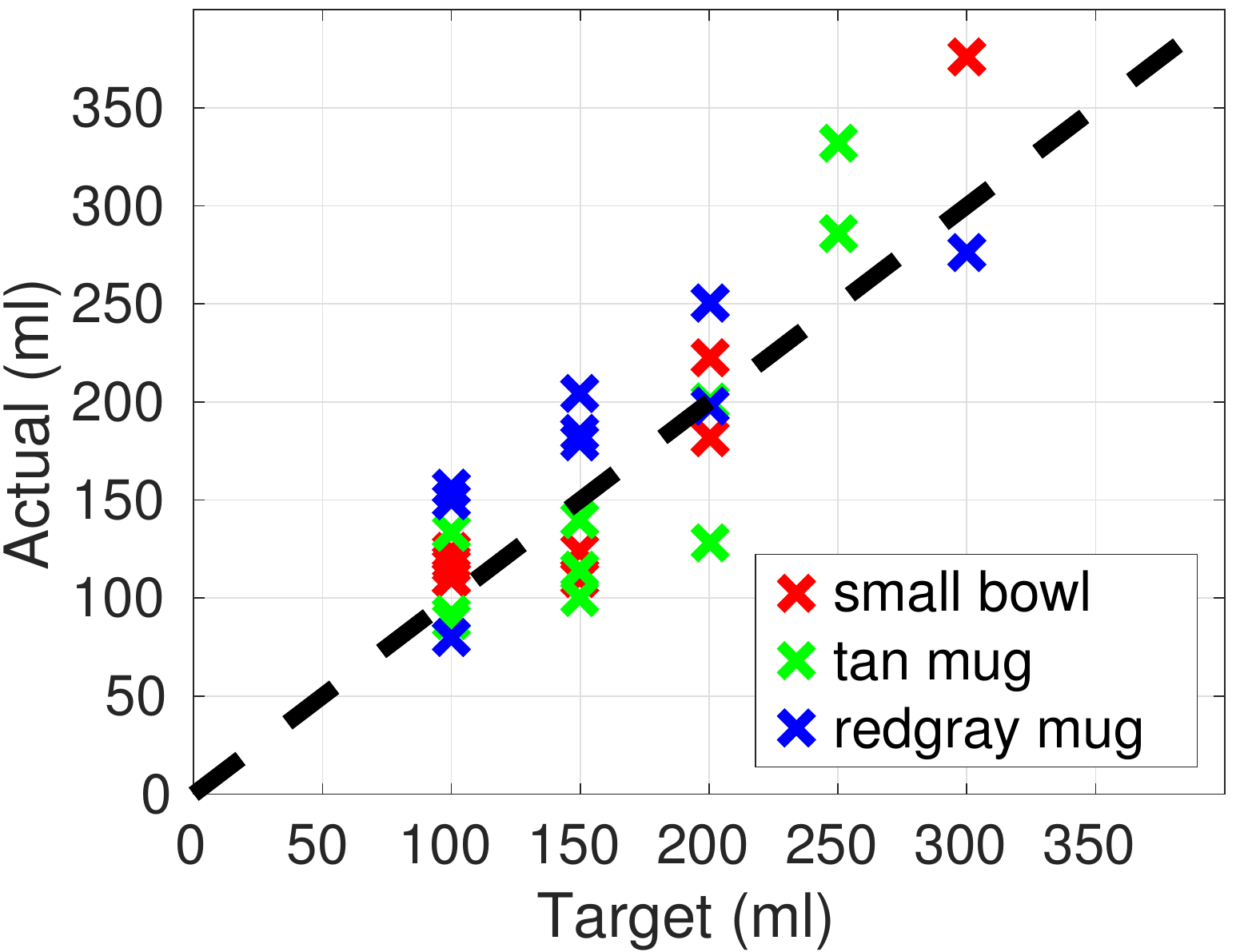}}
    \end{picture}
    \vspace{-0.1cm}
    \caption{Plot of the result of each pour using our model-free method as input to the controller. The x-axis is the target amount that the robot was attempting to reach, and the y-axis is the actual amount the robot poured. The points are color-coded by the target container. The black dashed line shows a 1:1 correspondence for reference.}
    \label{fig:control_points}
    \vspace{-0.3cm}
\end{figure}

\begin{figure}
    \centering
    \setlength{\fboxsep}{0pt}
    \setlength{\fboxrule}{1pt}
    \setlength{\unitlength}{1.0cm}
    \begin{subfigure}{2.5cm}
        \fbox{\includegraphics[width=2.5cm]{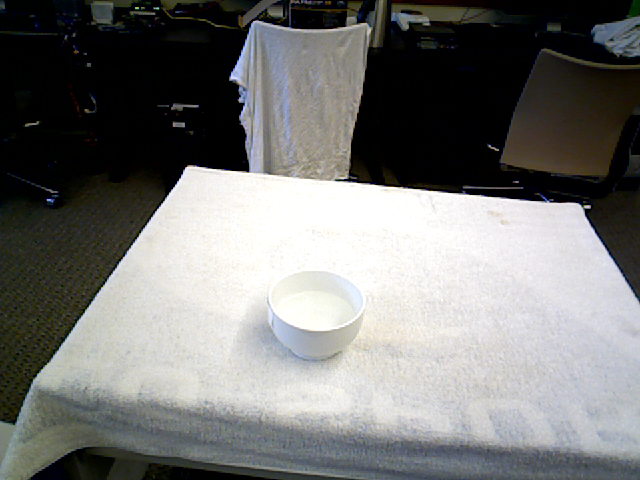}}
        \caption{200ml}
    \end{subfigure}\hspace{0.5cm}%
    \begin{subfigure}{2.5cm}
        \fbox{\includegraphics[width=2.5cm]{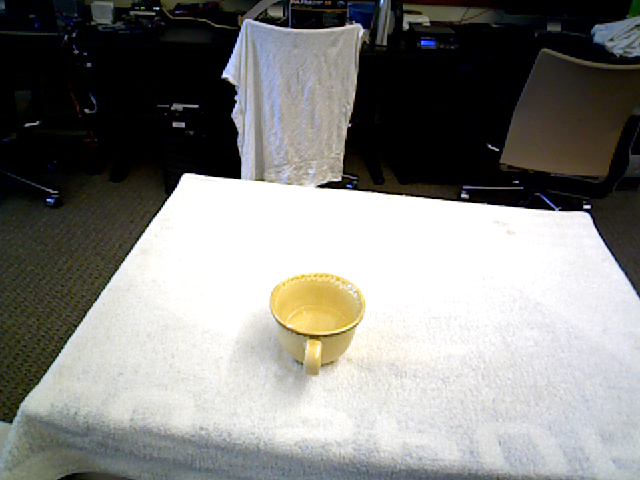}}
        \caption{100ml}
    \end{subfigure}\hspace{0.5cm}%
    \begin{subfigure}{2.5cm}
        \fbox{\includegraphics[width=2.5cm]{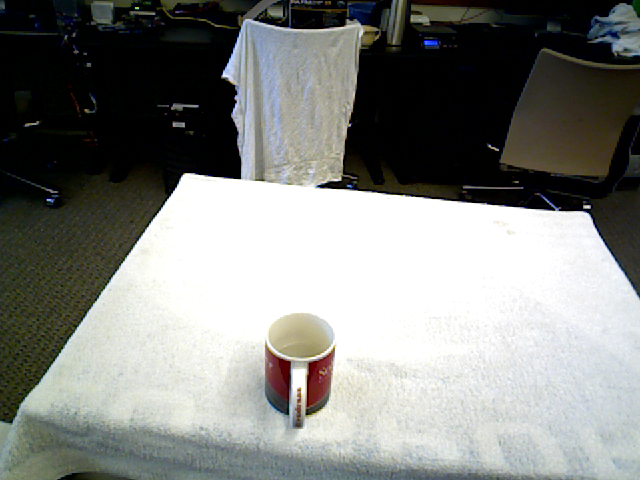}}
        \caption{150ml}
    \end{subfigure}

    \begin{subfigure}{2.5cm}
        \fbox{\includegraphics[width=2.5cm]{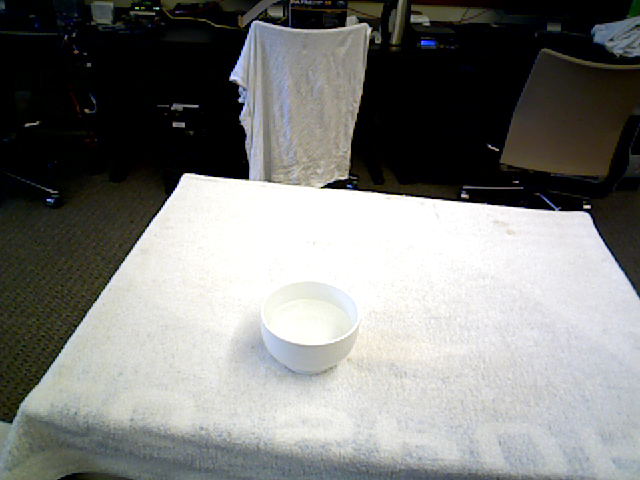}}
        \caption{250ml}
    \end{subfigure}\hspace{0.5cm}%
    \begin{subfigure}{2.5cm}
        \fbox{\includegraphics[width=2.5cm]{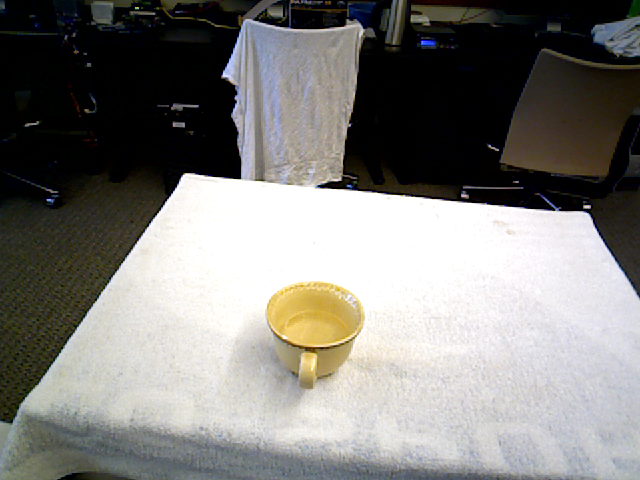}}
        \caption{150ml}
    \end{subfigure}\hspace{0.5cm}%
    \begin{subfigure}{2.5cm}
        \fbox{\includegraphics[width=2.5cm]{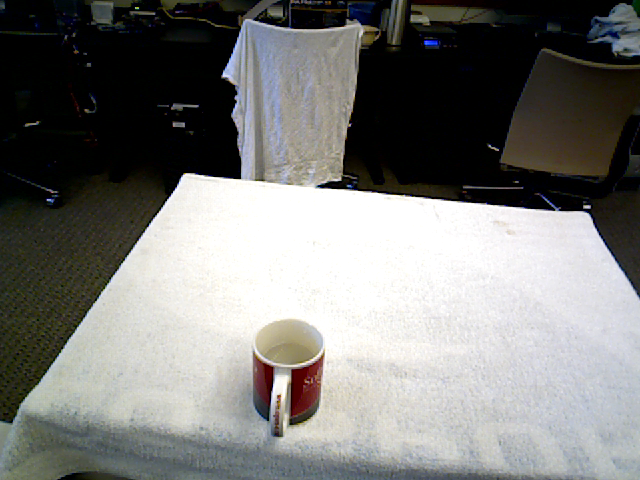}}
        \caption{200ml}
    \end{subfigure}
    \caption{Reference images for each of the three target containers. This is exactly the perspective the robot sees when looking at the containers. Notice that a 50ml difference is difficult to perceive even for a human (best viewed enlarged).}
    \label{fig:refrence_amounts}
    \vspace{-0.75cm}
\end{figure}

Estimating the volume {\it a posteriori} and using a volume estimator as input to a pouring controller are two very different problems.
A volume estimation method may work well analyzing the data after the pouring is finished, but that does not necessarily mean it is suitable for control.
For example, if the estimator outputs an erroneous value at one timestep, it may be able to correct in the next since the trajectory of the pour does not change.
However, if this happens during a pour and the estimator outputs an erroneous value, this may result in a negative feedback loop in which the trajectory deviates more and more from optimal, leading to more erroneous volume estimates, etc.
To verify that our chosen method from the previous section is actually suitable for control, we need to execute it on a real robot for real-time control.

Even though our volume estimation is rather accurate, it is not clear whether it is good enough for actually pouring certain amounts of liquid. 
This is due to the fact that erroneous estimates during a pouring operation can generate a feedback loop between the volume estimator and the controller, resulting in poor performance.  
Thus, to verify that our method is actually suitable for control, we need to execute it in real-time on a real robot.

To test the multi-frame network with detections only, we executed 30 pours on the real robot using the PID controller described in section \ref{sec:robot_controller}.
We ran 10 sequences on each of the three target containers.
For each sequence, we randomly selected a target volume in $\{100,150,200,250,300\}$ milliliters and we randomly initialized the volume of water in the source container as either 300, 350, or 400 milliliters, always ensuring at least a 100ml difference between the starting amount in the source and the target amount (so the robot cannot simply dump out the entire source and call it a success).
Each pour lasted exactly 25 seconds, and we evaluated the robot based on the actual amount of liquid in the target container (as measured by a scale) after the pour was finished.

Figure \ref{fig:control_points} shows a plot of each pour, where the x-axis is the target amount and the y-axis is the actual volume of liquid in the target container after the pour finished.
Note that the robot performs approximately the same on all containers.
This is particularly interesting since the volume estimation network is never given any information about the target container, and must simply infer it based on the motion of the liquid.
Additionally, almost all of the 30 pours were within 50ml of the target.
In fact, the average error over all the pours was 38ml.
For reference, Figure \ref{fig:refrence_amounts} shows 50ml differences for each of our 3 containers from the robot's perspective.
As is apparent from this figure, 50ml is a small amount, and a human solving the same task would be expected to have a similar error.

\vspace{-0.1cm}
\section{Conclusion and Future Work}
\vspace{-0.2cm}

In this paper, we introduce a framework for visual closed-loop control for
pouring specific amounts of liquid into a container.  To provide real-time
estimation of the amount of liquid poured so far, we develop a deep network
structure that first detects the presence of water in individual pixels of color
videos and then estimates the volume based on these detections. 
We show how to automatically generate the
data and labels required to train the deep networks using heated water along with a calibrated RGB-D / thermal camera system. 
A model-based approach allows us then to estimate the volume of liquid
in a container based on the pixel-level water detections.  

Our experiments indicate that the deep network architecture can be trained to
provide real-time estimates from color only data that are slightly better than
the model-based estimates using thermal imagery. Furthermore, once trained on
multiple containers, our volume estimator does not require a matched shape model
of the target container.  We incorporated our approach into a PID
controller and found that it on average only missed the target amount by 38ml.
While this is not accurate enough for some applications (e.g., some industrial
settings), it is well suited for similar pouring tasks in standard home
environments.  To our knowledge, this is the first work that has combined
visual feedback with control in order to pour specific amounts of liquids into
everyday containers.

This work opens up various directions for future research.  
One important avenue is to develop methods to improve the robustness of the neural networks.
Other interesting directions include generalization to different liquids, such as pouring a glass of soda or a cup of coffee, representing target amounts in \emph{relative} terms, such as in ``Pour me a half cup of water'', and more sophisticated control schemes on top of our perception.

\bibliographystyle{IEEEtran}
\vspace{-0.4cm}
{\footnotesize 
\bibliography{icra2017}}

\end{document}